\documentclass[10pt,twocolumn,letterpaper]{article}
\usepackage{iccv}
\usepackage{times}
\usepackage{epsfig}
\usepackage{graphicx}
\usepackage{amsmath}
\usepackage{amssymb}
\usepackage{subcaption}
\usepackage{epstopdf}
\usepackage{cases}
\usepackage{multirow}
\usepackage{ulem}
\usepackage{bbding}
\usepackage{makecell}
\usepackage{enumitem}

\usepackage[pagebackref=true,breaklinks=true,colorlinks,bookmarks=false]{hyperref}

\iccvfinalcopy 

\def\iccvPaperID{4895} 

\ificcvfinal\fi

\begin{document}

\title{Center Contrastive Loss for Metric Learning}

\author{Bolun Cai, Pengfei Xiong, Shangxuan Tian \\
Shopee\\
{\tt\small \{caibolun,xiongpengfei2019,tshxuan\}@gmail.com}
}

\maketitle
\ificcvfinal\thispagestyle{empty}\fi

\begin{abstract}

Contrastive learning is a major studied topic in metric learning. However, sampling effective contrastive pairs remains a challenge due to factors such as limited batch size, imbalanced data distribution, and the risk of overfitting. In this paper, we propose a novel metric learning function called \textbf{Center Contrastive Loss}, which maintains a class-wise center bank and compares the category centers with the query data points using a contrastive loss. The center bank is updated in real-time to boost model convergence without the need for well-designed sample mining. The category centers are well-optimized classification proxies to re-balance the supervisory signal of each class. Furthermore, the proposed loss combines the advantages of both contrastive and classification methods by reducing intra-class variations and enhancing inter-class differences to improve the discriminative power of embeddings. Our experimental results, as shown in Figure \ref{fig:recall}, demonstrate that a standard network (ResNet50) trained with our loss achieves state-of-the-art performance and faster convergence. The code will be released soon.

\end{abstract}

\section{Introduction}
\begin{figure}[ht!]
\centering  
\includegraphics[width=1.0\linewidth]{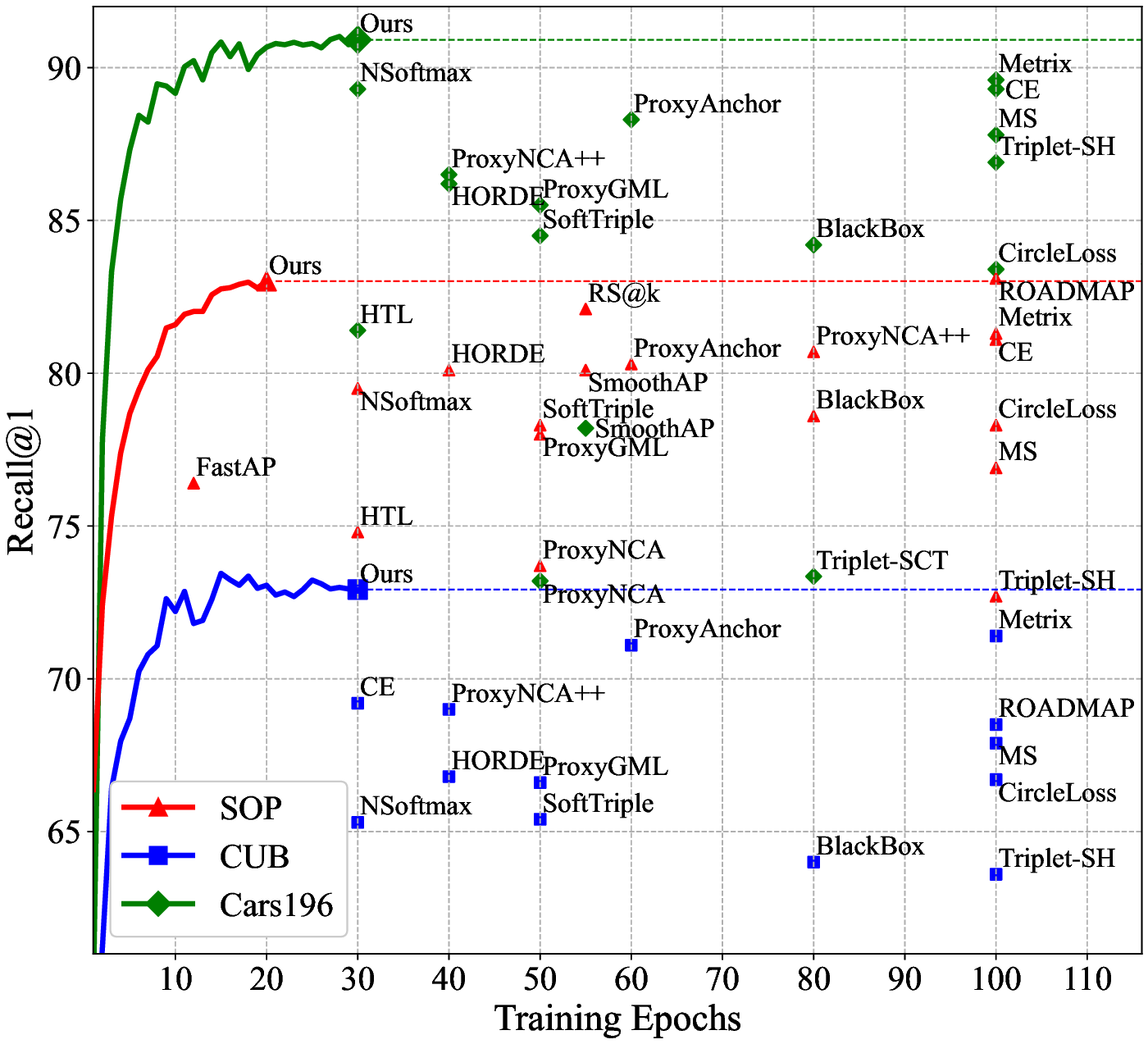} 
\caption{
Accuracy in Recall@1 versus training epochs on the SOP \cite{liftstruct}, CUB \cite{cub} and Cars196 \cite{car} dataset. Our loss achieves the highest accuracy and converges faster than the other methods, where the details of related methods are described in Section \ref{sec:experiments}.}  
\label{fig:recall}  
\end{figure}

Metric learning is widely used in computer vision to learn effective similarity measures from high-dimensional data, and it has been applied to various tasks such as image retrieval \cite{benchmark, liftstruct,ce}, person re-identification \cite{circleloss,mgn}, and face recognition \cite{facenet,centerloss,cosface}. Recently, deep neural networks have been successful in learning complex and nonlinear mappings to extract suitable embeddings. Contrastive learning \cite{contrast,liftstruct,xbm} has become a major research direction due to its success in representation learning. 

\begin{figure*}[ht!]
\centering
\subfloat[Lifted-struct \cite{liftstruct}]{\includegraphics[height=4.5cm]{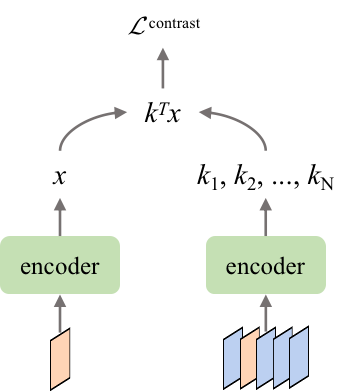}}\qquad
\subfloat[XBM \cite{xbm}]{\includegraphics[height=4.5cm]{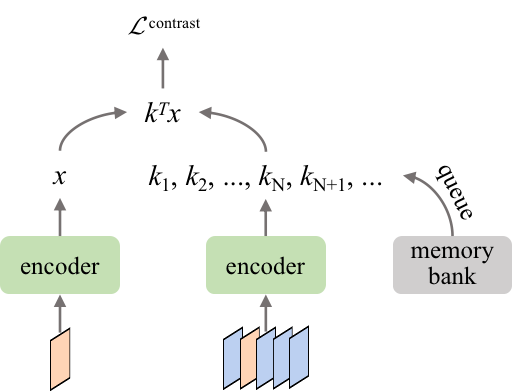}}\qquad
\subfloat[Ours]{\includegraphics[height=4.5cm]{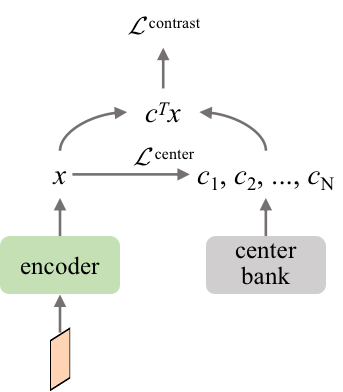}}
\caption{Conceptual comparison of difference contrastive mechanisms. (a) The encoder computes the query and contrasts it with one positive example and multiple, which is updated end-to-end \cite{simclr}. (b) The query data is contrasted with the embeddings sampled from a memory bank \cite{mb}, which is maintained as a queue with the mini-batches in the past iterations. (c) Our method maintains and updates a memory bank of category centers $\{c_j\}$ by $\mathcal{L}^{\mathrm{center}}$ in sync with the encoder, and contrasts them with the query data $x$ by $\mathcal{L}^{\mathrm{contrast}}$.
} 
\label{fig:framework}
\end{figure*}

The contrastive loss \cite{contrast} measures pairwise similarities between data points in the embedding space, where relevant pairs are pulled as close as possible, while irrelevant ones are pushed far apart. To capture more relational information beyond pairwise data, a group of multiple pairs \cite{facenet,npair,liftstruct,rll} is utilized to provide rich supervisory signals. For instance, lifted-struct \cite{liftstruct}, as shown in Figure \ref{fig:framework}(a), is a simple framework with an end-to-end contrastive mechanism, similar to SimCLR \cite{simclr}. It collects sufficient informative pairs from each batch, but low-quality pairs in the current batch may not contribute to training or may even hinder embedding learning. Therefore, various sampling techniques \cite{sct,strategy,fanng,sh,hdc} have been introduced to mine high-quality pairs. However, the hard mining ability is inherently limited by the batch size, and these techniques may reduce the generalization ability and increase the risk of overfitting. Moreover, using a large group of contrastive pairs can result in high training complexity and slow convergence.

To overcome the limitations of mining samples within a single batch, the cross-batch memory (XBM) \cite{xbm} introduces a memory bank \cite{mb} shown in Figure \ref{fig:framework}(b), which records the embeddings of recent iterations, allowing for mining informative samples across multiple batches. However, the embeddings in the memory bank are enqueued only when they were last seen, resulting in updating out-of-sync. To address this issue, a momentum encoder called MoCo is adopted in \cite{moco,mocov3}. In addition, these memory bank methods face long-tailed distribution problems in real-world applications: 1) for imbalanced datasets, high-frequency classes have a higher lower bound of loss and contribute much higher importance than low-frequency classes; 2) for large-class datasets, the limited size of the memory bank results in insufficient samples of each class to mine effective information. Later on, cluster contrastive methods \cite{spcl,prism,cluster_contrast} enforce the cluster assignments rather than comparing instance sampling to avoid the large memory bank or large batch size. However, these methods divide cluster updating and metric learning, which is discussed in Section \ref{sec:cluster}.

To overcome these limitations, we propose a novel loss function called \textbf{Center Contrastive Loss (CCL)}, as illustrated in Figure \ref{fig:framework}(c), which constrains a class-wise \textbf{center bank} updating in real-time and contrasts it with the data points by a \textbf{contrastive loss}. 
Compared to end-to-end mechanisms \cite{facenet,npair,liftstruct,rll}, the number of category centers is substantially smaller than the large group of contrastive pairs, boosting model convergence and robustness against noisy labels without the need for well-designed sample mining. 
Compared to memory bank mechanisms \cite{xbm,mb,moco,mocov3}, the category centers in the bank are updated in sync with the encoder, and re-balance the supervisory signal of each class suitable for imbalanced or large-class datasets.
Furthermore, the proposed loss leverages the advantages of both contrastive and classification methods. The center contrast provides well-optimized classification proxies in both the compact intra-class variations and separable inter-class differences.


Thanks to these advantages, our method achieves state-of-the-art Recall@1 accuracy on several commonly used datasets, such as Cars196 \cite{car}, SOP \cite{liftstruct} and CUB \cite{cub}, and it converges quickly with only one-fifth of training epoch comparing with previous state-of-the-art methods, \textit{e.g.} ROADMAP \cite{roadmap}, Metrix \cite{metrix}, as shown in Figure \ref{fig:recall}.




\section{Related Works}
\label{sec:related}

\begin{figure*}[ht!]
\centering
\subfloat[Contrastive \cite{contrast}]{\includegraphics[width=0.152\textwidth]{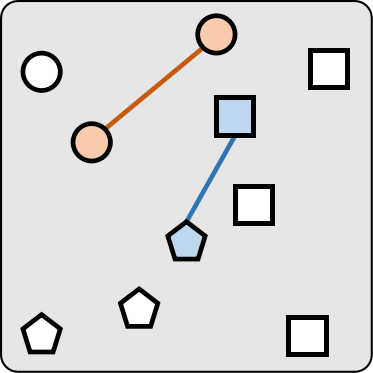}}\qquad
\subfloat[Triplet \cite{facenet}]{\includegraphics[width=0.152\textwidth]{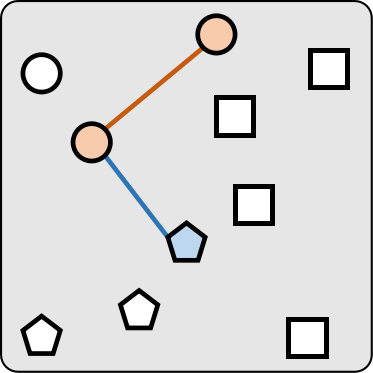}}\qquad
\subfloat[$N$-Pair \cite{npair}]{\includegraphics[width=0.152\textwidth]{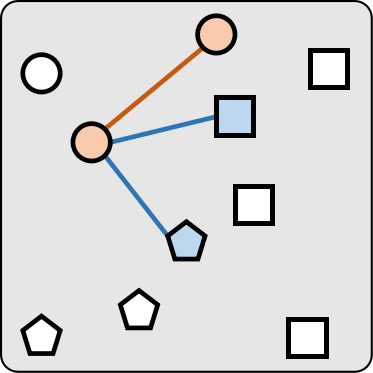}}\qquad
\subfloat[Lifted-Struct \cite{liftstruct}]{\includegraphics[width=0.152\textwidth]{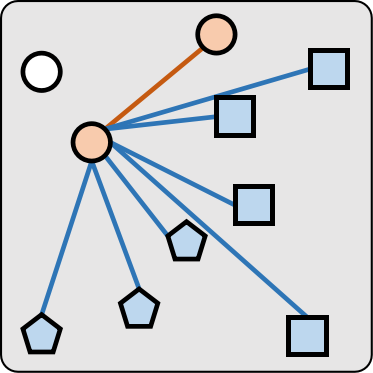}}\qquad
\subfloat[Ranked List \cite{rll}]{\includegraphics[width=0.152\textwidth]{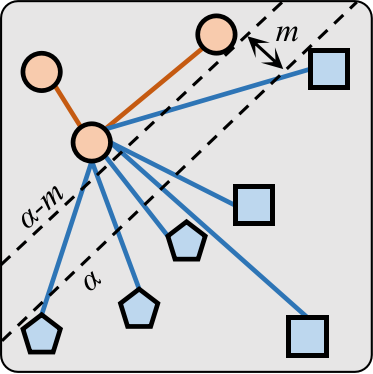}} \vskip+4px
\subfloat[ProxyNCA \cite{proxynca}]{\includegraphics[width=0.152\textwidth]{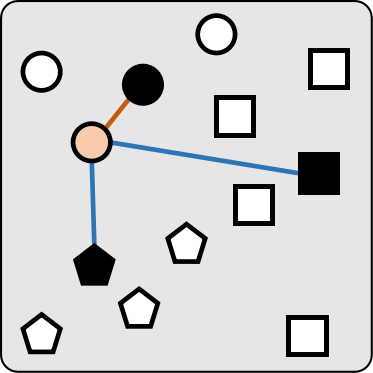}}\qquad
\subfloat[SoftTriple \cite{softtriple}]{\includegraphics[width=0.152\textwidth]{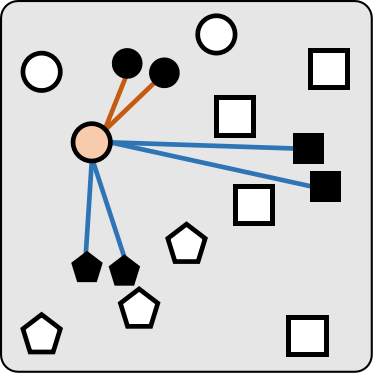}}\qquad
\subfloat[ProxyGML \cite{proxygml}]{\includegraphics[width=0.152\textwidth]{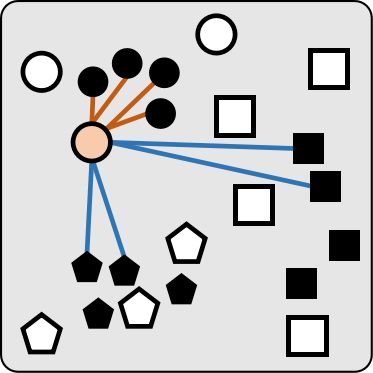}}\qquad
\subfloat[ProxyAnchor \cite{proxyanchor}]{\includegraphics[width=0.152\textwidth]{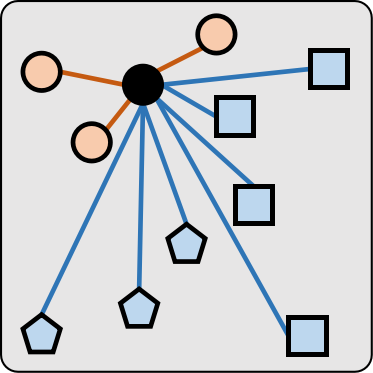}}\qquad
\subfloat[Ours]{\includegraphics[width=0.152\textwidth]{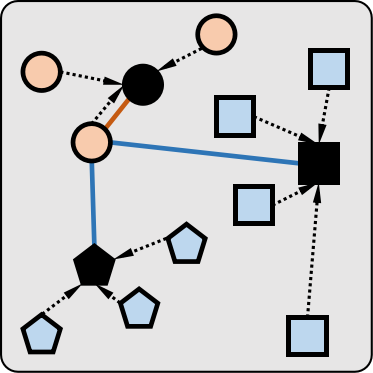}}
\caption{Illustration of popular metric learning losses and ours. Different shapes represent different categories, where \textbf{\textcolor[RGB]{212,81,0}{orange}}, \textbf{\textcolor[RGB]{0,119,187}{blue}}, and \textbf{black} shapes indicate positive data points, negative data points, and proxies, respectively. (a) Contrastive loss is trained on the distance between a pair of examples. (b) Triplet loss contrasts each data point with only one positive example and one negative example. (c) $N$-pair loss and (d) lifted-struct loss incorporate one positive example and multiple negative classes. (e) Ranked list loss not only exploits all negative examples but also all positive ones. (f) ProxyNCA loss associates each data point only with proxies. (g) SoftTriple loss and (h) ProxyGML loss assign multiple proxies to each category to reflect intra-class variance. (i) ProxyAnchor loss handles the entire data and associates them with each proxy. (j) Our proposed loss utilizes each category center as a classification proxy, which is explicitly optimized by all training data. For further details, please refer to the text.} 
\label{fig:losses}
\end{figure*}

In this section, the related metric learning approaches are reviewed for two families regarding the type of loss.

\subsection{Contrastive Losses}

\textbf{Pairwise losses} have been widely used in deep metric learning. As depicted in Figure \ref{fig:losses}(a), a contrastive loss \cite{contrast} was firstly introduced for this task, which pulls a pair of embeddings together if they have the same label and pushes them apart otherwise. Since image retrieval is a typical ranking task, recent pairwise losses aim to utilize higher order relations to improve feature mining. As shown in Figure \ref{fig:losses}(b-d), triplet loss \cite{facenet}, $N$-pair loss \cite{npair}, and lifted-struct loss associate a data point with single positive and multiple negative examples, where the negatives are pulled away considering their difficulty. In contrast, ranked list loss \cite{rll}, as shown in Figure \ref{fig:losses}(e), separates the positive and negative sets with a large margin to take into account all data. However, optimization over all pairs is impractical, so several sampling technologies \cite{sct,strategy,fanng,sh,hdc} have been proposed to find informative pairs or triplets.

\textbf{Listwise losses} has also been explored in deep metric learning to obtain differentiable rank approximations. In \cite{hl}, histogram loss is used to minimize the distribution of similarity between non-relevant pairs being larger than that of relevant ones. Based on information theory, RankMI loss \cite{rankmi} maximizes the mutual information between samples within the same semantic class. Recently, average precision (AP) \cite{ap}, as a standard retrieval evaluation metric, has been used as the optimization objective for listwise ranking. FastAP \cite{fastap} and SoftBin \cite{softbin} utilize smoothed discretization of similarity scores through soft-binning techniques to approximate the rank function. To overcome the brittleness of AP with respect to small score variations, a generic black box combinatorial solver \cite{blackbox} is introduced for AP optimization. Other approaches rely on explicitly approximating the non-differentiable rank functions (\textit{e.g.} SoDeep \cite{sodeep}), or with a sum of sigmoid functions in the recent SmoothAP approach \cite{smoothap}. While these methods provide elegant AP upper bounds, they are generally coarse AP approximations. In addition, large batch size is crucial for ranking loss, which is often limited by hardware constraints.

\subsection{Classification Losses}

\textbf{Large-margin losses.} In \cite{nsoftmax,ce}, it was shown that the standard classification loss, cross-entropy loss, serves as a strong baseline for deep metric learning. To improve embedding discrimination, large-margin losses have been widely applied in the domain of face retrieval. Liu \textit{et al.} \cite{lsoftmax} proposed a large-margin softmax (L-Softmax) loss by adding multiplicative angular to constrain each identity. SphereFace \cite{sphereface} and additive angular softmax (AM-Softmax) loss \cite{amsoftmax} further improved the L-Softmax loss by normalizing the weights. To overcome the optimization difficulty of SphereFace, CosFace \cite{cosface} and ArcFace \cite{arcface} moved the angular margin into the cosine and arc-cosine space, respectively. In our proposed loss, we also applied the general technology of large margin to further improve performance.

\textbf{Proxy-based losses} provide another variant of classification loss. The first proxy-based loss is ProxyNCA \cite{proxynca} (Figure \ref{fig:losses}(f)), which is an approximation of neighborhood component analysis (NCA) using proxies. To reflect intra-class variance, SoftTriple \cite{softtriple} and ProxyGML \cite{proxygml} extend a single proxy to multiple proxies for each class, as illustrated in Figure \ref{fig:losses}(g-h) respectively. ProxyAnchor loss \cite{proxyanchor}, as shown in Figure \ref{fig:losses}(i), associates the entire data point and each proxy with consideration of their relative hardness determined by data-to-data relations. However, the proxies as a part of the trainable parameters are only optimized by the relative relations in a batch. In our proposed loss, center constraint explicitly provides well-optimized proxies by all training data to find the global category centers. 

\section{Method}\label{sec:method}

To address the inherent limitations of previous methods, we propose a novel metric learning loss called \textbf{Center Contrastive Loss}, which maintains \textit{a list} of category centers as \textit{classification proxies} and compares them with the query data point by a \textit{contrastive loss} with \textit{large-margin}. In this section, we first review the InfoNCE loss \cite{infonce}, a representative contrastive loss. Then, we infer and analyze the proposed loss in detail.

\subsection{Review of Contrastive Loss}

Contrastive learning \cite{contrast} is a well-established framework that learns effective representations from data organized into similar and dissimilar pairs. Recently, several studies \cite{simclr,mb,moco,mocov3} have presented promising results in visual representation learning by using approaches related to contrastive loss. Given a query data point $x$ and a set of contrastive samples $\{k_1, k_2, k_3, ...\}$, the contrastive loss is a function whose value is low when $x$ is similar to its positive sample $k_+$ and dissimilar to all the others $\{k_{-}\}$. When the similarity is measured by the dot product $k^T x$ between $\ell$-2 normalized $x=\tilde{x} /\| \tilde{x} \|_2$ and $k=\tilde{k} / \| \tilde{k}\|_2$, a form of a contrastive loss function called InfoNCE \cite{infonce} is considered in this paper:
\begin{align}
\mathcal{L}^{\mathrm{contrast}} &= -\log\dfrac{e^{k_+^T x /\tau}}{\sum _{\{k\}}{e^{k^T x/\tau}}} \notag \\
& = -\log\dfrac{e^{k_+^T x /\tau}}{e^{k_+^T x /\tau} +\sum_{\{k_-\}}e^{k_-^T x /\tau}}, \label{eq:infonce}
\end{align}
where $\tau$ denotes a temperature parameter \cite{temperature}. 

The contrastive loss is computed across all sample pairs, both $(x, k_+)$ and $(x, k_-)$, in a mini-batch. For lifted-struct contrastive learning \cite{liftstruct}, the sum is over one positive $k_+$ and $N-1$ negative samples $\{k_-\}$, as shown in Figure \ref{fig:framework}(a). To mine informative samples across multiple batches, the cross-batch memory \cite{xbm} (Figure \ref{fig:framework}(b)) contrasts each query $x$ with a memory bank maintained as a queue with the current batch enqueued and the oldest dequeued.

\subsection{Center Contrastive Loss}

A memory bank is composed of the embeddings of all samples from the previous epoch and cannot be updated in sync with the encoder. Additionally, when the number of classes is large, the samples of each class in the memory bank may are not enough to extract effective information. To address these issues, we propose a center bank that maintains and updates category centers in real-time as the contrastive samples, as shown in Figure \ref{fig:framework}(c). Moreover, the center constraint minimizes the intra-class variations to enhance the discriminative power of the embeddings.

To this end, $c_y$ denotes the $y$-th class center of embeddings and replaces the contrastive samples $k$ in Eq. \eqref{eq:infonce} with $k_+=c_y$ and $k_-=c_{j\ne y}$.
\begin{numcases}{}
\mathcal{L} ^{\mathrm{contrast} } = -log\dfrac{e^{c_y^T x /\tau}}{e^{c_y^T x /\tau} +\sum_{j\ne y}e^{c_j^T  x /\tau}} \label{eq:contrast}\\
\mathcal{L} ^{\mathrm{center} } = \left \| x - c_y \right \| ^2 \label{eq:center}
\end{numcases}
Here, $c_j$ is updated as the embeddings change with the center loss function Eq. \eqref{eq:center}. The center vector $c_y$ is also $\ell$-2 normalized. After normalizing, the transformed vectors $x$ and $c_y$ have unit norms and
\begin{equation}
\| x-c_y \| ^2=\| x \|_2^2+\| c_y \|_2^2-2c_y^Tx=-2c_y^Tx+2.
\label{eq:norm_center}
\end{equation}
Therefore, the minimization of $\| x-c_y \| ^2$ is equivalent to the maximization of $c_y^Tx$, and the class center optimized in Euclidean space is equivalent to the one in cosine space.

In metric learning, contrastive losses with large-margin can further reinforce the optimization. L-Softmax \cite{lsoftmax} adds angular constraints to improve feature discrimination, and A-Softmax \cite{sphereface} improves L-Softmax by normalizing the weights. Due to the non-monotonicity of the multiplicative margin, the decision boundary of them is difficult to be optimized. To address this problem, CosFace \cite{cosface} defines a cosine measure margin $m$ with the decision boundary given by $c_y^Tx-m$, which is applied in this paper. Here, the learned embeddings are distributed on a hypersphere, and the reciprocal of the temperature parameter $\tau$ can be regarded as the hypersphere radius $s=1/\tau$. Since the query $x$ with label $y$ is distributed around each center $c_y$ on a hypersphere, a similarity penalty $m$ is employed between $x$ and $c_y$, which simultaneously enhances the intra-class compactness and inter-class discrepancy. Finally, we adopt the joint supervision of contrastive loss with large-margin and center loss to train the model, which is given as follows:
\begin{align}
\mathcal{L} = &\mathcal{L} ^{\mathrm{contrast} } + \lambda \mathcal{L} ^{\mathrm{center} } \notag \\
= &-\log\dfrac{e^{s\cdot(c_y^Tx-m)+2\lambda\cdot c_y^Tx}}{e^{s\cdot(c_y^Tx-m)} +\sum_{j\ne y}e^{s\cdot c_j^Tx}}, \label{eq:ours}
\end{align}
where $\lambda$ is a scalar used for loss balancing.

\subsection{Analysis}

\begin{figure*}[ht!]
\centering
\subfloat[Cross-entropy \cite{ce}]{\includegraphics[width=0.22\textwidth]{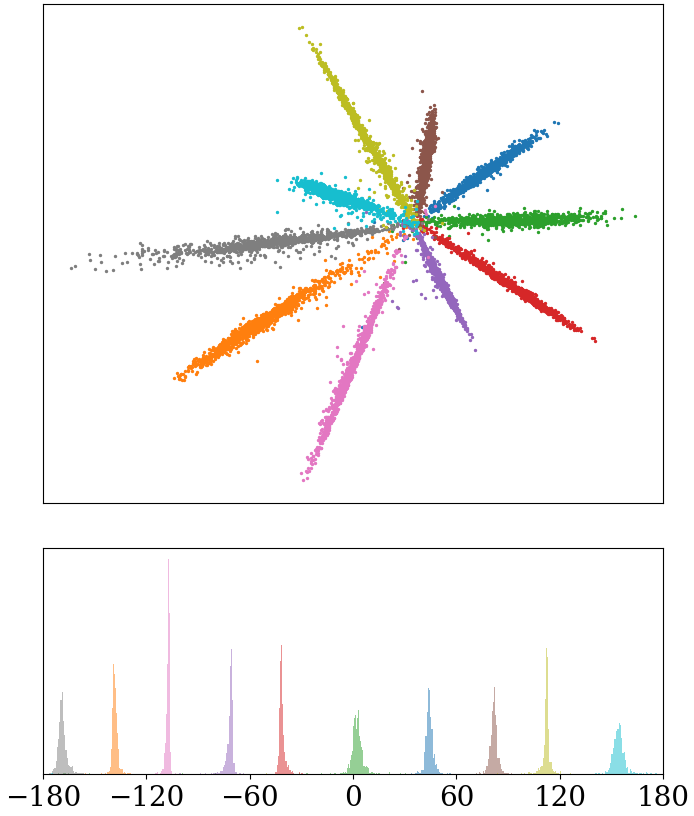}}\quad
\subfloat[NSoftmax \cite{nsoftmax}]{\includegraphics[width=0.22\textwidth]{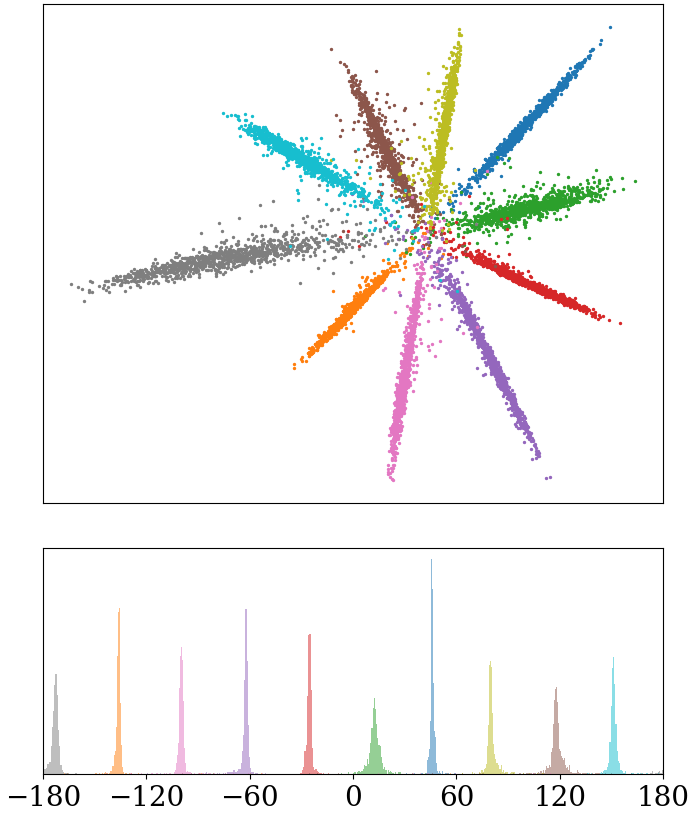}}\quad
\subfloat[Center Loss \cite{centerloss}]{\includegraphics[width=0.22\textwidth]{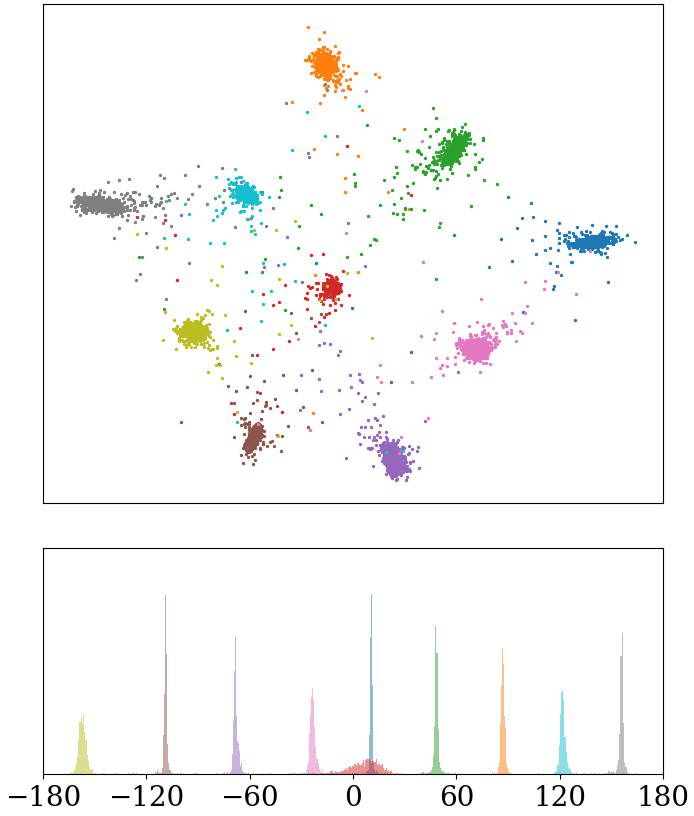}}\quad
\subfloat[Ours]{\includegraphics[width=0.22\textwidth]{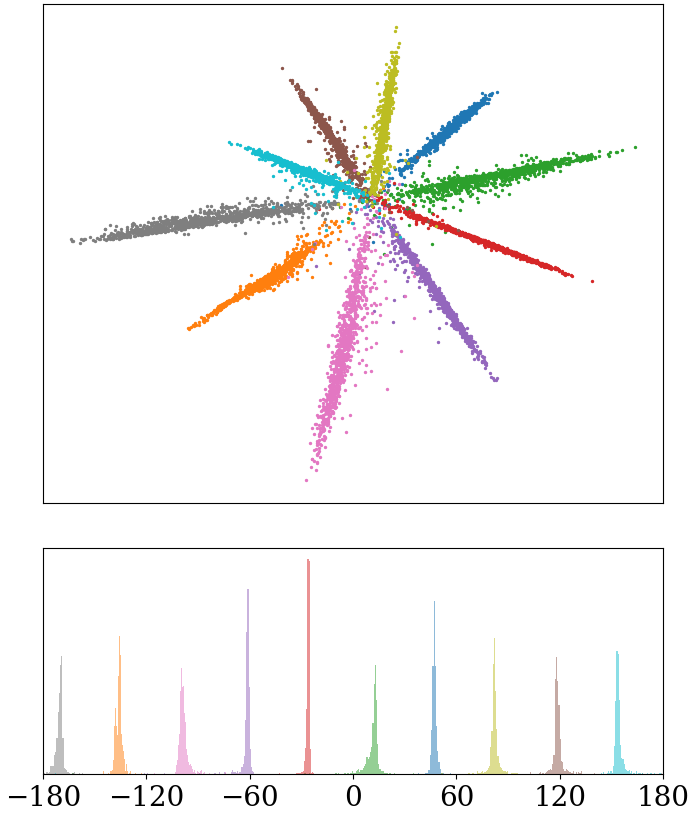}}\quad
\caption{The distribution of embedding under difference losses. We reduced the output number of the hidden layer in LeNet to 2 and trained it on MNIST \cite{lenet} for 10 epochs using AdamW optimizer \cite{adamw} with a fixed learning rate (using default parameters in \textit{Pytorch}). The first row shows the distance distribution in Euclidean space, and the second row shows the angular distribution in cosine space. Focusing on the center constraint, the embeddings in (d) are supervised by our loss without large margin ($m=0$ and $\lambda=2.0$). Compared to NSoftmax loss in (b), the radial bandwidth in Euclidean space is more narrower, and the angular distribution in cosine space is more cohesive. Additionally, compared to the center loss in (c), the embeddings are distinguishable in both Euclidean and cosine spaces. Best viewed in color.
} 
\label{fig:mnist}
\end{figure*}

In this subsection, we compare the proposed method to cross-entropy \cite{ce}, center loss \cite{centerloss}, NSoftmax \cite{nsoftmax}, ProxyNCA \cite{proxynca}, large-margin \cite{cosface}, and cluster contrastive \cite{spcl,prism,cluster_contrast}.

\subsubsection{Comparison to Cross-entropy}\label{sec:ce}
Without the center constraint ($\lambda=0$), the proposed loss degenerates to cross-entropy loss by replacing cosine similarity with a fully connected layer $c^Tx=W^T\tilde{x}+b$ and setting $s=1$, $m=0$. The standard cross-entropy loss is given by:
\begin{equation}
\mathcal{L} ^{\mathrm{CE}}  = -\log\dfrac{e^{W^T_y\tilde{x} +b_y}}{\sum_{j}e^{W^T_j\tilde{x} +b_j}}.
\end{equation}
Here, $W_j$ denotes the $j$-th column of the weights in the fully connected layer, and $b$ is the bias term, where $W$ and $\tilde{x}$ are without $\ell$-2 normalization. The last fully connected layer is a linear classifier, and the deep features of different classes are distinguished by the decision boundary of $\|W\|\cos(\theta)$. As shown in Fig. \ref{fig:mnist}(a), cross-entropy loss tends to create a radial feature distribution, but the no-normalization features and bias terms result that the embeddings cannot be distributed elegantly on the hypersphere.

\subsubsection{Comparison to Center Loss}\label{sec:centerloss}
Based on the standard cross-entropy loss, additional parameters $c_j$ are introduced as the center constraint to minimize intra-class variations while keeping the embeddings of different classes separable, as formulated below:
\begin{equation}
\mathcal{L} ^{\mathrm{CenterLoss}} = -\log\dfrac{e^{W^T_y \cdot \tilde{x} +b_y}}{\sum_{j}e^{W^T_j \cdot \tilde{x} +b_j}} + \lambda \left \| \tilde{x} - c_y \right \| ^2
\end{equation}
Due to the unconformity between the entropy measure of cross-entropy loss and the Euclidean measure of the center constraint, the retrieval embeddings have to be transformed by principal component analysis (PCA) and compared under cosine measure in \cite{centerloss}. As shown in Fig. \ref{fig:mnist}(c), the embeddings are discriminative within a wide range in Euclidean space but confused by vector angles in cosine space. From the perspective of contrastive learning, our proposed loss blends the center constraint and linear classifier ($W_j=c_j$) under a unified cosine measure ($x$ and $c_j$ are $\ell$-2 normalized). Moreover, the unified parameters and measure of joint supervision Eq. \eqref{eq:ours} have a consistent optimization direction, making the model converge quickly.

\subsubsection{Comparison to NSoftmax}\label{sec:nsoftmax}
When training the model with cross-entropy, the normalized softmax (NSoftmax) loss \cite{nsoftmax} removes the bias term $b$ in the last fully connected layer and adds an $\ell$-2 normalization module to the inputs $\tilde{x}$ and weights $W$ to optimize the cosine similarity. In addition, a hypersphere radius $s$ is used to exaggerate the difference among classes and boost the gradients. 
\begin{equation}
\mathcal{L} ^{\mathrm{NSoftmax}} = -\log\dfrac{e^{s\cdot c_y^Tx}}{\sum_{j}e^{s\cdot c_j^Tx}},
\end{equation}
which results in a decision boundary given by: $\cos(\theta)$. The key difference and advantage of our proposed loss over NSoftmax is the center constraint between the data and weights of each class. As illustrated in Fig. \ref{fig:mnist}(c), we can see that, by radial variations and angular distribution, the NSoftmax loss can perfectly classify samples in the cosine space. However, it is not quite robust to noise because there is without intra-class constraint -- any small perturbation around the decision boundary can change the decision.

\subsubsection{Comparison to ProxyNCA}\label{sec:proxynca}
If the class weights $c_y$ are viewed as proxies, the NSoftmax loss can be classified as one of the proxy-based losses. When $\lambda=0$, $m=0$, and the term $e^{s\cdot c_y^Tx}$ in the denominator of Eq. \eqref{eq:ours} equaling to zero, our proposed loss is converted into ProxyNCA loss:
\begin{equation}
\mathcal{L}^{\mathrm{ProxyNCA}} = -log\dfrac{e^{s\cdot c_y^Tx}}{\sum_{j\ne y}e^{s\cdot c_j^Tx}}.
\label{eq:proxynca}
\end{equation}
Here, the slight difference in the denominator affects the optimization direction, which we analyze as follows. For simplicity, we denote the similarity measure as $\mathcal{S}(x,c) = s\cdot c^Tx$, and the contrastive loss $\mathcal{L}^{\mathrm{contrat}}$ can be rewritten as follows:
\begin{align}
\mathcal{L}^{\mathrm{contrat}} & = -\log \frac{e^{\mathcal{S}(x,c_y)}}{e^{\mathcal{S}(x,c_y)}+\sum_{j\ne y}{e^{\mathcal{S}(x,c_j)}}} \notag\\
& = \log(1+\textstyle \sum_{j\ne y}{e^{\mathcal{S}(x,c_j)-\mathcal{S}(x,c_y)}}) \label{eq:rewrite2}
\end{align}
The gradient of Eq. \eqref{eq:rewrite2} with respect to $\mathcal{S}(x,c)$ is given by:
\begin{equation}
\frac{\partial \mathcal{L}^{\mathrm{contrast} }}{\partial \mathcal{S}(x,c)} = \begin{cases}
\dfrac{-\textstyle \sum_{j\ne y}{e^{\mathcal{S}(x,c_j)-\mathcal{S}(x,c)}}}{1+\textstyle \sum_{j\ne y}{e^{\mathcal{S}(x,c_j)-\mathcal{S}(x,c)}}}, &c=c_y  \\
\dfrac{e^{\mathcal{S}(x,c)-\mathcal{S}(x,c_y)}}{1+\textstyle \sum_{j\ne y}{e^{\mathcal{S}(x,c_j)-\mathcal{S}(x,c_y)}}}, &c\ne c_y
\end{cases}. \notag
\end{equation}
In the same way, the gradient of \eqref{eq:proxynca} can be expressed as:
\begin{equation}
\frac{\partial \mathcal{L}^{\mathrm{ProxyNCA} }}{\partial \mathcal{S}(x,c)} = \begin{cases}
-1, &c=c_y  \\
\dfrac{e^{\mathcal{S}(x,c)}}{\textstyle \sum_{j\ne y}{e^{\mathcal{S}(x,c_j)}}}, &c\ne c_y
\end{cases}. \notag
\end{equation}
In the ProxyNCA loss, the scale of the gradient is constant for every positive example, which damages the flexibility and generalizability of embedding learning. Conversely, the gradient of $\mathcal{L}^{contrast}$ for a positive example ($c=c_y$) is correlated with the contrastive relation:
\begin{itemize}[itemsep=2pt,topsep=0pt,parsep=0pt]
\item If the query is close to the positive center ($c_y^Tx\to 1$) and far from negative centers ($c_{j\ne y}^Tx\to 0$), $(\mathcal{S}(x,c_j)-\mathcal{S}(x,c))$ tends to $-s \ll 0$. As a result, $\sum_{j\ne y} \exp(\mathcal{S} (x,c_j)-\mathcal{S} (x,c_y))\to 0$, and the gradient tends to zero.
\item Conversely, when $c_y^Tx\to 0$ and $c_{j\ne y}^Tx\to 1$, we have $(\mathcal{S}(x,c_j)-\mathcal{S}(x,c))$ tending to $s\gg 0$. Therefore, $\sum_{j\ne y} \exp(\mathcal{S} (x,c_j)-\mathcal{S} (x,c_y))\to \infty $ and $ \partial \mathcal{L}^{\mathrm{contrast}}$ approximates to $-1$. 
\end{itemize}

\subsubsection{Comparison to Large-margin}\label{sec:margin}
To improve the decision boundary of the NSoftmax loss, large-margin technology introduces an additive similarity margin $m$. Based on Eq. \eqref{eq:rewrite2}, the contrastive loss with large margin can be defined as follows:
\begin{equation}
\mathcal{L}^{\mathrm{contrat}} = \log(1+\textstyle \sum_{j\ne y}{e^{\mathcal{S}(x,c_j)-(\mathcal{S}(x,c_y)-m)}}).
\label{eq:margin}
\end{equation}
The large-margin loss has a similar but not identical optimization goal to the center constraint $\mathcal{L}^{\mathrm{center}}$. Here, we provide an understanding in terms of $\mathcal{L}^{\mathrm{contrat}}$. As $\exp(\Delta)$ is a monotonically increasing function and $\exp(\Delta)$ is always positive, so we have the inequation as follows:
\begin{align}
\exp(\max(\{\Delta\}_{j=1}^{N}))\le \textstyle\sum_{j}^{N}{\exp(\Delta_j)}\le & \notag\\ 
N\exp(\max(\{\Delta\}_{j=1}^{N})),&
\end{align}
where $\Delta_y=0$ and $\Delta_{j\ne y}=\mathcal{S}(x,c_{j\ne y})-\mathcal{S}(x,c_y)+m$. Then, we apply $\log(\cdot)$ to obtain:
\begin{align}
\max(\{0, \Delta_1, ..., \Delta_N\}) \le \log(1+\textstyle\sum_{j\ne y}{\exp(\Delta_j)})&\le \notag \\ 
\max(\{0, \Delta_1,..., \Delta_N\})+\log(N).&
\label{eq:inequation}
\end{align}
Therefore, Eq. \eqref{eq:margin} can be treated as a differentiable approximation to the maximum of $\mathcal{S}(x,c_{j\ne y})-\mathcal{S}(x,c_y)+m$. During the training, the model seeks to minimize the maximum among all the positives and negatives. The optimization goal of large-margin loss is that all the entries $\mathcal{S}(x,c_y)-\mathcal{S}(x,c_{j\ne y})\ge m$, which is equivalent to $\mathcal{S}(x,c_y)\ge \mathcal{S}(x,c_{j\ne y})+m$. According to Eq. \eqref{eq:norm_center}, the optimization goal of center constraint $\mathcal{L}^{\mathrm{center}}$ is $\mathcal{S}(x,c_y)=1$, which is the recall goal (Recall@1) of image retrieval and the extreme case of large-margin loss ($\mathcal{S}(x,c_{j\ne y})=0$, $m=1$). Therefore, the large-margin and center constraint can jointly improve the performance, as verified in Section \ref{sec:parameter}.

\subsubsection{Comparison to Cluster Contrastive}\label{sec:cluster} 
In \cite{spcl,prism,cluster_contrast}, the $y$-th cluster center is momentum updated by the query features belonging to class $y$ in the mini-batch as:
\begin{equation}
c_y = \mu \cdot c_y + (1-\mu)x_y,
\label{eq:update}
\end{equation}
where $\mu$ is a momentum coefficient. The updating direction of momentum average is same as the gradient of Eq. \eqref{eq:center}. However, the updating direction of our method is not only impacted by Eq. \eqref{eq:center} to pull the samples in the same cluster as close as possible, but also is optimized by Eq. \eqref{eq:contrast} to push the different cluster center far apart. When the gradient of $c_y$ does not back-propagate which is a constant in the contrastive subproblem of Eq. \eqref{eq:contrast}, our proposed method is equivalent to cluster contrastive methods. Shown in Table \ref{tab:stop_grad}, the proposed loss is significantly better than cluster contrastive with stop-gradient operation, especially on product retrieval dataset (SOP and InShop).
\begin{table}[ht]
\centering
\begin{tabular}{l|cccc}
\hline
Method      & SOP & CUB & Cars196 & InShop \\
\hline
CCL w/ stop-grad & 69.8 & 72.1 & 90.5 & 80.8 \\
CCL         & 83.1 & 73.5 & 91.0 & 92.3 \\
\hline
\end{tabular}
\caption{Accuracy in Recall@1 compared to cluster contrastive.}
\label{tab:stop_grad}
\end{table}

\section{Experiments}
\label{sec:experiments}

In this section, we present an evaluation of our proposed method and compare it to the state-of-the-art methods on four benchmark datasets \cite{liftstruct,cub,cars,inshop}. Additionally, we investigate the effect of hyperparameters ($m$ and $\lambda$) and embedding dimensionality to demonstrate the robustness of our method. Our implementation uses the \textit{PyTorch} library\footnote{Our code is modified and adapted on \cite{ce}: \url{https://github.com/jeromerony/dml_cross_entropy/}.} and initializes the ResNet50 \cite{resnet} model with weights pre-trained on ImageNet \cite{mocov3}. 

\subsection{Datasets}

There are four commonly used datasets for evaluating metric learning: Stanford Online Product (SOP) \cite{liftstruct}, Caltech-UCSD Birds-200-2011 (CUB) \cite{cub}, Cars196 \cite{cars}, and In-shop Clothes Retrieval (InShop) \cite{inshop}. For SOP \cite{liftstruct}, the standard retrieval split is followed, where 59,551 images from 11,318 classes are used for training and 60,502 images from the remaining classes are used for testing. For CUB \cite{cub}, the model is trained on 5,864 images from the first 100 classes and evaluated on 5,924 images from the rest of the classes. Similarly, for Cars196 \cite{cars}, 8,054 images from the first 98 classes are used for training, while 8,131 images from the remaining classes are kept for testing. As for InShop \cite{inshop}, the benchmark setting is followed, where 25,882 images from the first 3,997 classes are used for training and 28,760 images from the remaining classes are used for testing, where the test set is further partitioned into a query set with 14,218 images from 3,985 classes and a gallery set with 12,612 images from 3,985 classes.

\subsection{Implementation Details}

In our experiments, we employ the common random sampling method among all samples, with a mini-batch size of 128 for all experiments, as in most classification training schemes. During training, input images are augmented by random cropping and horizontal flipping, while they are center-cropped during testing. To compare our results to those of HORDE \cite{horde}, ProxyAnchor \cite{proxyanchor}, ROADMAP \cite{roadmap}, and Triplet-SCT \cite{sct}, we implement models trained and tested with $256\times256$ cropped images. For CUB and Cars196, we found that random jittering of the brightness, contrast, and saturation slightly improves the results.

In all experiments, we train the models using SGD with Nesterov acceleration \cite{nesterov} and a weight decay of 0.0005. For SOP and InShop, we set the learning rate to 0.003 and 0.006, respectively, with a momentum of 0.99. For CUB and Cars196, the learning rate is set to 0.02 and 0.05 without momentum. To reduce overfitting, we use label smoothing \cite{inception} for the target probabilities. Following \cite{ce}, we set the target of positive probability $e^{s\cdot c_y^Tx}$ to $1-\epsilon$, and the probabilities of the others $e^{s\cdot c_{j\ne y}^Tx}$ to $\epsilon/(N-1)$ (where $N$ is the number of centers) with $\epsilon = 0.1$ in all our experiments. Due to the relatively small number of training samples in CUB and Cars196, we freeze all the batch normalization layers in the feature encoder and add dropout with a probability of 0.2 before the loss function to further reduce overfitting. All of the implementation details can be found in the publicly available code.
\begin{table*}[ht]
\centering
\begin{tabular}{c|l|l|ccc|ccc|ccc|ccc}
\hline
    \multicolumn{2}{c|}{\multirow{2}*{Method}}&\multirow{2}*{Arch.}&\multicolumn{3}{c|}{SOP \cite{liftstruct}}&\multicolumn{3}{c|}{CUB \cite{cub}}&\multicolumn{3}{c|}{Cars196 \cite{cars}}&\multicolumn{3}{c}{InShop \cite{inshop}}\\\cline{4-15}
    \multicolumn{2}{c|}{}&&1&10&100&1&2&4&1&2&4&1&10&20\\
\hline
    \multirow{7}*{\rotatebox{90}{Pairwise losses}}
    &Triplet-SH \cite{sh}                   &R$^{512}$&72.7&86.2&93.8&63.6&74.4&83.1&86.9&92.7&95.6& -  & -  & -  \\
    &HTL \cite{htl}                 &I$^{512}$&74.8&88.3&94.8&57.1&68.8&78.7&78.8&87.0&92.2&80.9&94.3&95.8\\
    &MS \cite{ms}                   &I$^{512}$&78.2&90.5&96.0&65.7&77.0&86.3&84.1&90.4&94.0&89.7&97.9&98.5\\
    &CircleLoss \cite{circleloss}   &R$^{512}$&78.3&90.5&96.1&66.7&77.4&86.2&83.4&89.8&94.1&-&-&-\\
    &XBM \cite{xbm}                 &I$^{512}$&79.5&90.8&96.1&65.8&75.9&84.0& -  & -  & -  &89.9&97.6&98.4\\
    &HORDE$^\dagger$ \cite{horde}   &I$^{512}$&80.1&91.3&96.2&66.8&77.4&85.1&86.2&91.9&95.1&90.4&97.8&98.4\\
    &Triplet-SCT$^\dagger$ \cite{sct}                 &R$^{512}$&81.9&92.6&96.8&57.7&69.8&79.6&73.4&82.0&88.0&90.9&97.5&98.1\\
\hline
    \multirow{9}*{\rotatebox{90}{Classification losses}}
    &ProxyNCA \cite{proxynca}       &R$^{512}$&73.7&-&-&49.2&61.9&67.9&73.2&82.4&86.4&-&-&-\\
    &CosFace \cite{cosface,benchmark}&R$^{512}$&75.8&-&-&67.3&-&-&85.5&-&-&-&-&-\\
    &ArcFace \cite{arcface,benchmark}&R$^{512}$&76.2&-&-&67.5&-&-&85.4&-&-&-&-&-\\
    &ProxyGML \cite{proxygml}       &I$^{512}$&78.0&90.6&96.2&66.6&77.6&86.4&85.5&91.8&95.3&-&-&-\\
    &SoftTriple \cite{softtriple}   &R$^{512}$&78.3&90.3&95.9&65.4&76.4&84.5&83.2&89.5&94.0&-&-&-\\
    &NSoftmax \cite{nsoftmax}       &R$^{2048}$&79.5&91.5&96.7&65.3&76.7&85.4&89.3&94.1&96.4&89.4&97.8&98.7\\
    &ProxyNCA++ \cite{proxynca_pp}  &R$^{512}$&80.7&92.0&96.7&69.0&79.8&87.3&86.5&92.5&95.7&90.4&98.1&98.8\\
    &ProxyAnchor$^\dagger$ \cite{proxyanchor}      &I$^{512}$&80.3&91.4&96.4&71.1&80.4&87.4&88.3&93.1&95.7&91.9&98.1&98.7\\
    &CE \cite{ce}                   &R$^{2048}$&81.1&91.7&96.3&69.2&79.2&86.9&89.3&93.9&96.6&90.6&98.0&98.6\\
    &Metrix \cite{metrix}   &R$^{512}$&81.3&92.7&97.1&70.4&80.6&\textcolor{red}{\textbf{88.7}}&88.5&93.4&96.5&91.9&98.1&98.8\\
\hline
    \multirow{7}*{\rotatebox{90}{Listwise losses}}
    &RLL \cite{rll} &I$^{512}$&76.1&89.1&95.4&57.4&69.7&79.2&74.0&83.6&90.1&-&-&-\\
    &FastAP \cite{fastap}           &R$^{512}$&76.4&89.0&95.1&-&-&-&-&-&-&90.9&97.7&98.5\\
    &BlackBox \cite{blackbox}       &R$^{512}$&78.6&90.5&96.0&64.0&75.3&84.1&84.2&90.4&94.4&88.1&97.0&97.9\\
    &SmoothAP \cite{smoothap}       &R$^{512}$&80.1&91.5&96.6&-&-&-&76.1&84.3&89.8&-&-&-\\
    &SoftBin \cite{softbin}         &R$^{512}$&80.6&91.3&96.1&61.2&73.1&83.0&-&-&-&-&-&-\\
    &RS@k \cite{rsk}                &R$^{512}$&82.1&92.8&97.0&-&-&-&88.2&93.0&95.9&-&-&-\\
    &ROADMAP$^\dagger$ \cite{roadmap} &R$^{512}$&\textcolor{red}{\textbf{83.1}}&92.7&96.3&68.5&78.7&86.6&-&-&-&-&-&-\\
\hline
    \multirow{5}*{\rotatebox{90}{Ours$^\dagger$}}
    &Baseline  ($m,\lambda=0$) &R$^{512}$&80.8&92.5&97.3&71.1&80.8&\textcolor{blue}{\underline{88.2}}&87.7&92.4&95.7&90.2&98.0&98.7\\
    &&&\small\textcolor{green}{+1.5}&\small\textcolor{green}{+0.5}&\small\textcolor{green}{+0.1}&\small\textcolor{green}{+0.7}&\small\textcolor{green}{+0.0}&\small\textcolor{green}{-0.4}&\small\textcolor{green}{+1.9}&\small\textcolor{green}{+1.5}&\small\textcolor{green}{+0.7}&\small\textcolor{green}{+1.9}&\small\textcolor{green}{+0.4}&\small\textcolor{green}{+0.2}\\
    &CCL \ \ ($m=0$)                       &R$^{512}$&\textcolor{blue}{\underline{82.3}}&\textcolor{blue}{\underline{93.0}}&\textcolor{blue}{\underline{97.4}}&\textcolor{blue}{\underline{71.8}}&\textcolor{blue}{\underline{80.8}}&87.8&\textcolor{blue}{\underline{89.6}}&\textcolor{blue}{\underline{93.9}}&\textcolor{blue}{\underline{96.4}}&\textcolor{blue}{\underline{92.1}}&\textcolor{blue}{\underline{98.4}}&\textcolor{blue}{\underline{98.9}}\\
    &&&\small\textcolor{green}{+2.3}&\small\textcolor{green}{+0.8}&\small\textcolor{green}{+0.1}&\small\textcolor{green}{+2.4}&\small\textcolor{green}{+1.1}&\small\textcolor{green}{-0.4}&\small\textcolor{green}{+2.3}&\small\textcolor{green}{+2.1}&\small\textcolor{green}{+1.1}&\small\textcolor{green}{+2.1}&\small\textcolor{green}{+0.5}&\small\textcolor{green}{+0.3}\\
    &CCL$^{*}$ ($m\ne0$)                    &R$^{512}$ &\textcolor{red}{\textbf{83.1}}&\textcolor{red}{\textbf{93.3}}&\textcolor{red}{\textbf{97.4}}&\textcolor{red}{\textbf{73.5}}&\textcolor{red}{\textbf{81.9}}&87.8&\textcolor{red}{\textbf{91.0}}&\textcolor{red}{\textbf{94.5}}&\textcolor{red}{\textbf{96.8}}&\textcolor{red}{\textbf{92.3}}&\textcolor{red}{\textbf{98.5}}&\textcolor{red}{\textbf{99.0}}\\
\hline   
\end{tabular}
\caption{Comparison with the state-of-the-art methods. Backbone architecture (Arch.) with superscripts denoting embedding sizes are denoted by abbreviations: I -- Inception \cite{inception}, R -- ResNet50 \cite{resnet}. $^\dagger$ indicates models using larger input images, and ``CCL$^*$" indicates our loss with large-margin. The \textcolor{red}{\textbf{blod}} indicates the best value while \textcolor{blue}{\underline{underline}} indicates the second best.}
\label{tab:results}
\end{table*}

\subsection{Comparison to Other Methods}

We compare our method to other state-of-the-art methods across four image retrieval datasets and report the results in Table \ref{tab:results}. We focus on recent methods with embedding dimension larger than 512 and divide them into three categories: pairwise losses \cite{sh,htl,ms,circleloss,xbm,horde,sct}, classification losses \cite{ce,nsoftmax,proxynca,cosface,arcface,proxygml,softtriple,proxynca_pp,proxyanchor,metrix}, and listwise losses \cite{rll,fastap,blackbox,smoothap,softbin,rsk,roadmap}. Our proposed method is a special classification loss that utilizes contrast between query and a list of category centers.

When $\lambda=0$ and $m=0$, our proposed method is equivalent to NSoftmax \cite{nsoftmax}, which is a strong baseline for deep metric learning. The performance improvement between our implementation and \cite{nsoftmax} mainly comes from the training tricks introduced in \cite{ce}. Based on this baseline, the center constraint (CCL, $\lambda \ne 0$ and $m = 0$) consistently improves Recall@k accuracy on all datasets with an increase of approximately 1.5, 0.7, 1.9, and 1.9 points on Recall@1, respectively. To further enhance the performance, we apply large-margin consistent with most other methods \cite{horde,proxyanchor,rll,roadmap}. Our loss with large-margin (CCL$^{*}$, $\lambda \ne 0$ and $m \ne 0$) can reinforce the generalization of embeddings, and results in a sustained increment of 2.3, 2.4, 2.3, and 2.1. As shown in Table \ref{tab:results}, our results with large-margin establish a new state-of-the-art across all methods for all datasets.

\begin{figure}[t]
\centering
\includegraphics[width=0.9\linewidth]{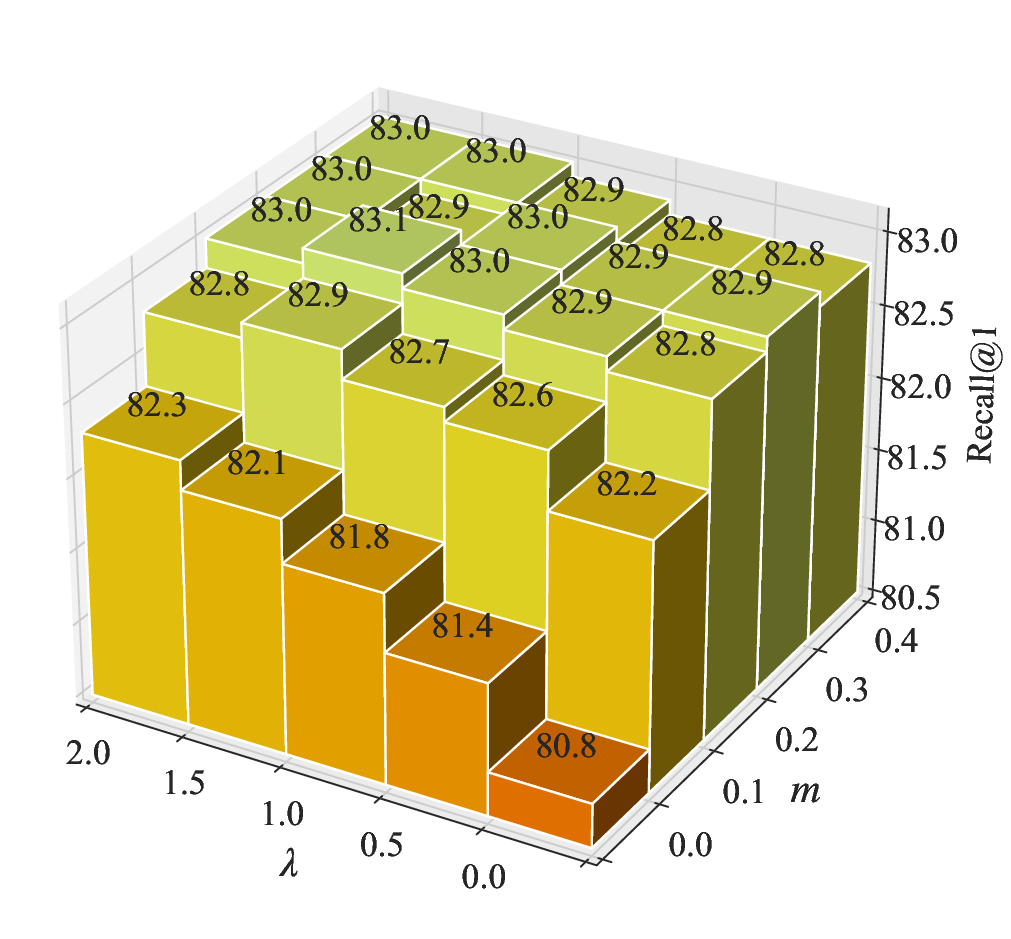}  
\caption{Accuracy in Recall@1 versus $\lambda$ and $m$ on the SOP dataset \cite{liftstruct}.}  
\label{fig:hist}
\end{figure}
\begin{figure}[t]
\centering
\includegraphics[width=0.95\linewidth]{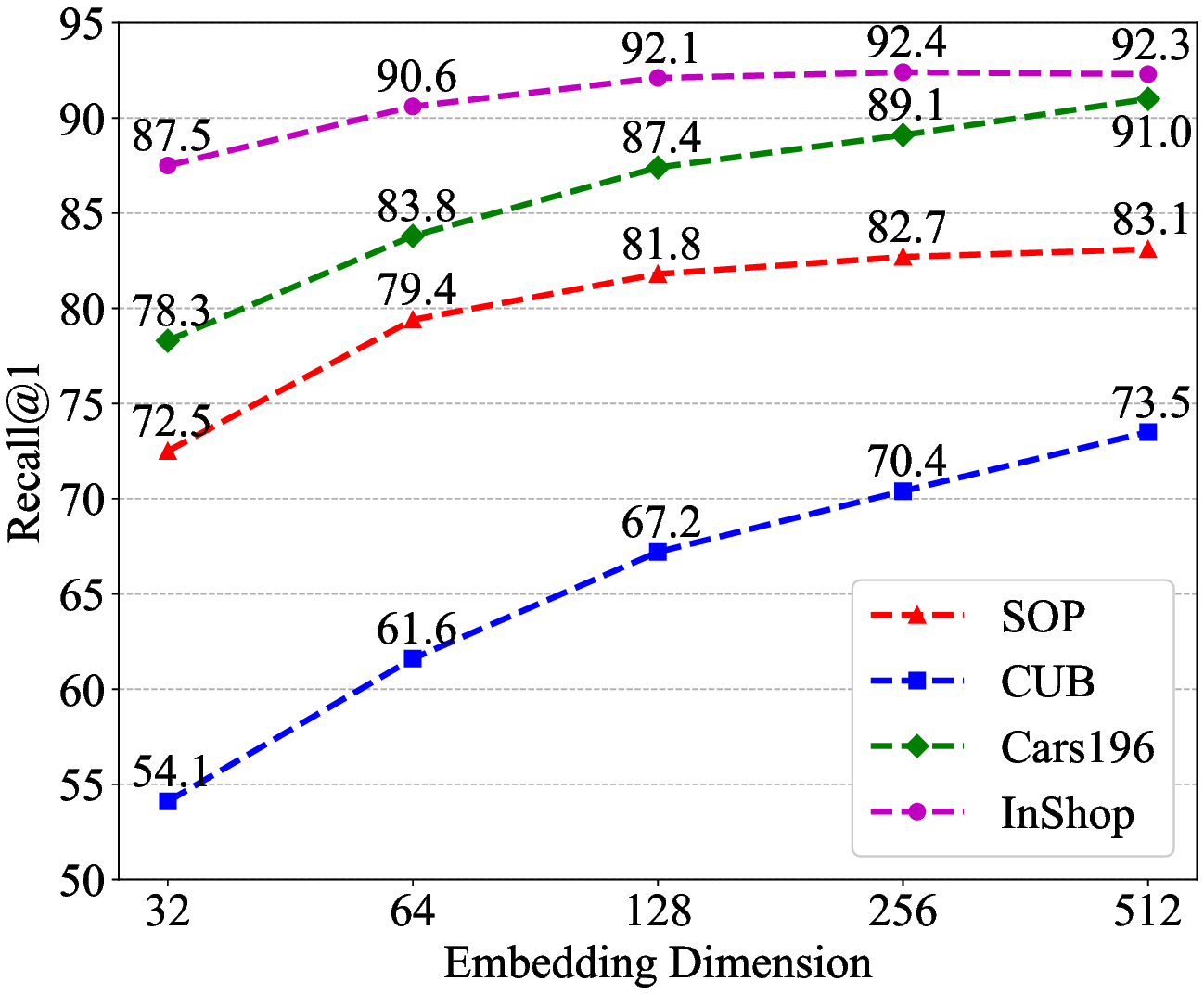}  
\caption{Accuracy in Recall@1 versus embedding dimensions.}  
\label{fig:dim}
\end{figure}

\subsection{Robustness Evaluation}
In the experiments, we adopt three types of noisy label: 1) symmetric noise, 2) long-tail noise and 3) real-world noise. Symmetric noise has been widely used to evaluate the model robustness \cite{coteaching,coteaching_plus}, where the symmetric noise is assigned to all classes with equal probability without regarding the similarity between data samples. To mimic the naturally occurring label noise, a long-tail noise proposed in \cite{prism} creates an openset label noise scenario as the ground-truth classes are eliminated in the corrupted dataset, where ground-truth class is randomly selected into a large number of small clusters. For real-world label noise, we use a Cars98N dataset \cite{prism} for training, and the test set of Cars196 is used for performance evaluation. 
Again following \cite{prism}, we use Inception \cite{inception} as the backbone architecture for all algorithms. Table \ref{tab:noise} shows the evaluation results on CUB and Cars196 under difference noisy label. To individually verify center contrastive for  noise robustness, our proposed method without large-margin ($m=0$) and with consistent center constraint ($\lambda=2.0$) achieves the highest performance among all the compared algorithms.

\begin{table*}[ht]
\centering
\begin{tabular}{l|cccc|ccccc}
\hline
    &\multicolumn{4}{c|}{CUB \cite{cub}}&\multicolumn{5}{c}{Cars196 \cite{cars}}\\\cline{2-10}
    \multirow{2}*{Noisy Rate}&\multicolumn{2}{c}{symmetric}&\multicolumn{2}{c|}{long-tail}&\multicolumn{2}{c}{symmetric}&\multicolumn{2}{c}{long-tail}&real-world\\
    &10\%&20\%&25\%&50\%&10\%&20\%&25\%&50\%&Cars98N \cite{prism}\\
\hline
    \multicolumn{10}{l}{\textit{Metric learning under label noise}}\\ \hline
    CircleLoss \cite{circleloss}&47.48&45.32&44.07&22.96&71.00&56.24&53.03&19.95&-\\
    ProxyNCA \cite{proxynca}&47.13&46.64&42.07&36.48&69.79&70.31&69.50&58.34&53.55\\
    Contrastive \cite{contrast} &51.77&51.50&47.27&39.43&72.34&70.93&65.60&26.45&44.91\\
    NSoftmax \cite{nsoftmax}&51.99&49.66&49.61&41.78&72.72&70.10&71.61&62.29&-\\
    FastAP \cite{fastap}&54.10&53.70&52.18&48.46&66.74&66.39&62.49&53.07&-\\
    MS \cite{ms} &57.44&54.52&53.60&41.66&66.31&67.14&63.92&43.73&49.00\\
    SoftTriple \cite{softtriple}&51.94&49.14&51.94&49.14&76.18&71.82&73.26&66.66&63.36\\
    XBM \cite{xbm}&56.72&50.74&52.25&41.58&74.22&69.17&69.46&36.43&38.73\\
\hline
    \multicolumn{10}{l}{\textit{Robust learning under label noise}}\\ \hline
    F-correction \cite{fcorrection} &53.41&52.60&-&-&71.00&69.47&-&-&-\\
    Co-teaching \cite{coteaching}&53.74&51.12&51.75&48.85&73.47&70.39&70.57&62.91&58.74\\
    Co-teaching+ \cite{coteaching_plus}&53.31&51.04&51.55&47.62&71.49&69.62&70.05&61.58&58.74\\
    Co-teaching w/ Temperature \cite{nsoftmax} &55.25&54.18&54.59&48.32&77.51&76.30&75.26&66.19&60.72\\
    SoftTriple + PRISM \cite{prism} &-&-&\textcolor{blue}{\underline{57.61}}&\textcolor{blue}{\underline{54.27}}&-&-&\textcolor{blue}{\underline{77.60}}&\textcolor{blue}{\underline{70.45}}&\textcolor{blue}{\underline{64.81}}\\
    XBM + PRISM \cite{prism} &\textcolor{blue}{\underline{58.78}}&\textcolor{blue}{\underline{58.73}}&55.77&53.46&\textcolor{blue}{\underline{80.06}}&\textcolor{blue}{\underline{78.03}}&77.08&68.26&57.95\\
\hline
CCL ($m=0$, $\lambda=2.0$) &\textcolor{red}{\textbf{62.34}}&\textcolor{red}{\textbf{62.69}}&\textcolor{red}{\textbf{61.88}}&\textcolor{red}{\textbf{58.73}}&\textcolor{red}{\textbf{81.17}}&\textcolor{red}{\textbf{78.53}}&\textcolor{red}{\textbf{78.10}}&\textcolor{red}{\textbf{70.90}}&\textcolor{red}{\textbf{67.57}}\\
\hline
\end{tabular}
\caption{Accuracy in Recall@1 on the CUB and Cars196 dataset with label noise. The \textcolor{red}{\textbf{blod}} indicates the best value while \textcolor{blue}{\underline{underline}} indicates the second best.}
\label{tab:noise}
\end{table*}

\subsection{Impact of Hyperparameters}
There are three hyperparameters in our proposed loss, including the hypersphere radius $s$, balancing scalar $\lambda$, and margin parameter $m$. Previous works \cite{proxyanchor,cosface,circleloss,mocov3} have shown that when $s$ is greater than $\sim$16, the performance of contrastive learning is high and stable, so we set $s$ to 16 in all experiments. In this study, we focus on investigating the effect of the two hyperparameters $\lambda$ and $m$. Additionally, we also investigate the robustness of our method to different embedding dimensions.

\subsubsection{$\lambda$ and $m$ of our loss}
\label{sec:parameter}

We conducted an analysis to investigate the impact of the hyperparameters $\lambda$ and $m$ in Eq. \eqref{eq:ours} on the SOP \cite{liftstruct} dataset. The results of this analysis are summarized in Figure \ref{fig:hist}, where we examine Recall@1 accuracy by varying the values of $m \in \{0.0, 0.1, 0.2, 0.3, 0.4\}$ and $\lambda \in \{0.0, 0.5, 1.0, 1.5, 2.0\}$. The balancing scalar $\lambda$ determines the intensity of the center constraint $\mathcal{L}^{\mathrm{center}}$. Without large margin ($m=0$), the accuracy in Recall@1 steadily improved from 80.8 to 82.3 as $\lambda$ increased, which verifies the effectiveness of our loss. Due to the strong representative learning provided by large-margin, increasing $\lambda$ improves performance although its effect is relatively small when $m$ is large. With the large-margin, the center constraint can still steadily improve the performance from 82.8 to 83.1. Therefore, he large-margin and center constraint can jointly improve the performance.

\subsubsection{Embedding dimension}
\label{sec:dimension}

The dimension of the embeddings is a crucial factor that affects the trade-off between speed and accuracy in image retrieval systems. Thus, we investigate the effect of embedding dimensions on Recall@1 accuracy. We test our loss with embedding dimensions varying from 32 to 512, as shown in Figure \ref{fig:dim}. The performance of our loss is fairly stable when the dimension is equal to or larger than 128. Surprisingly, our results with low embedding dimensions (\textit{e.g.} 256) are also competitive with previous methods with high dimensions shown in Table \ref{tab:results}.

\section{Conclusion}
\label{sec:conclusion}
In this paper, we propose a novel contrastive loss function called the center contrastive loss, which maintains and real-time updates a class-wise center bank to compare the category centers with the query data points by a contrastive loss. Our method provides well-optimized classification proxies and re-balances the supervisory signal of each class, combining the benefits of both contrastive and classification methods. As a result, our method achieves state-of-the-art performance on four public benchmark datasets and converges quickly without requiring careful sampling techniques. In the future, we plan to explore extensions of our loss for a wider range of applications, such as face recognition, person re-identification, and clustering.



{\small
\normalem
\bibliographystyle{ieee_fullname}
\bibliography{egbib}
}

\end{document}